\documentclass[letterpaper, 10pt, journal, twoside]{IEEEtran} 
\usepackage{caption}
\usepackage{subcaption}
\usepackage{times}
\usepackage{epsfig}
\usepackage{graphicx}
\usepackage{amsmath}
\usepackage{amssymb}
\usepackage{enumitem}
\DeclareMathOperator*{\argmin}{arg\,min}
\usepackage{algorithm}
\usepackage{algorithmic}
\usepackage{prettyref}
\usepackage{wrapfig}
\usepackage{hyperref}
\usepackage{xcolor}

\ifCLASSOPTIONcompsoc
  \usepackage[nocompress]{cite}
\else
  \usepackage{cite}
\fi

\begin{document}
\newcommand{\bl}[1]{{\color{blue}#1}}

\title{Sim-to-Real Brush Manipulation using Behavior Cloning and Reinforcement Learning}

\author{Biao~Jia,
        Dinesh~Manocha
\IEEEcompsocitemizethanks{\IEEEcompsocthanksitem B. Jia is with the Department
of Computer Science, University of Maryland at College Park,
MD, 20740.\protect\\
E-mail: \{biao, dm\}@cs.umd.edu
\IEEEcompsocthanksitem D. Manocha is with the Departments
of Computer Science and Electrical \& Computer Engineering, University of Maryland at College Park,
MD, 20740.}
}

\maketitle

\begin{abstract}
Developing proficient brush manipulation capabilities in real-world scenarios is a complex and challenging endeavor, with wide-ranging applications in fields such as art, robotics, and digital design. In this study, we introduce an approach designed to bridge the gap between simulated environments and real-world brush manipulation. Our framework leverages behavior cloning and reinforcement learning to train a painting agent, seamlessly integrating it into both virtual and real-world environments. Additionally, we employ a real painting environment featuring a robotic arm and brush, mirroring the MyPaint virtual environment. Our results underscore the agent's effectiveness in acquiring policies for high-dimensional continuous action spaces, facilitating the smooth transfer of brush manipulation techniques from simulation to practical, real-world applications.
\end{abstract}

\section{Introduction}
\begin{figure}[t!]
\centering
\includegraphics[width=0.45\textwidth]{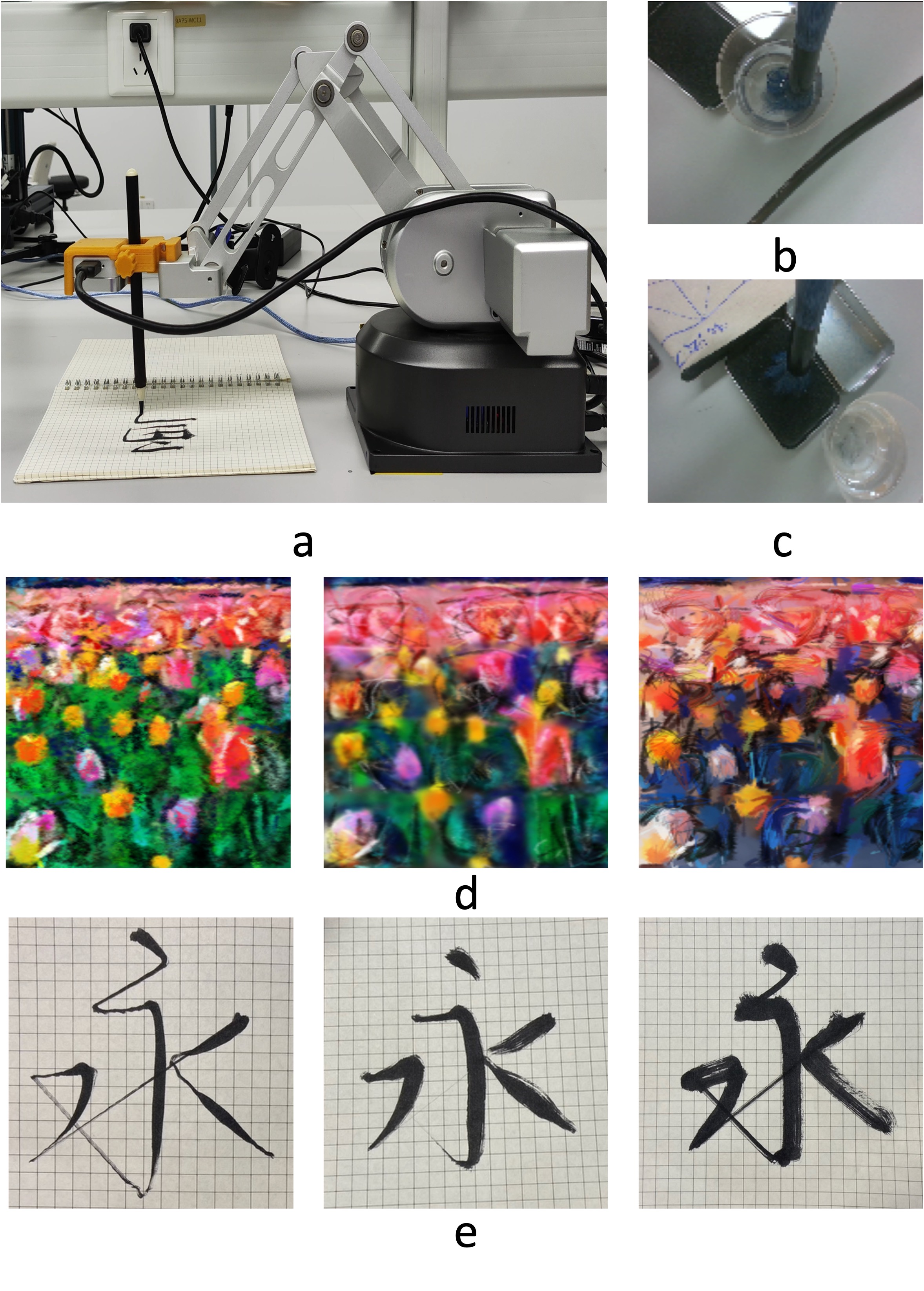}
\caption{\textbf{Demonstration of the Learned Model's Artistic Versatility}, showcasing a wide range of artistic styles achieved through real robot painting (\textbf{a: Robot setup with Realsense D415, UltraArm robot, and paintbrush; b: Water pot; c: Ink pot from the egocentric view from the camera mounted on the end-effector}) and digital painting outputs (\textbf{d: Painting in simulated environment; e: Painting using setup (a) with various brushes within the MyPaint [1] virtual environment}). Reference image courtesy of KanjiVG [2]. This demonstrates the effective transferability of the painting policy to a real environment, enabling the generation of various artistic styles.}
\label{fig:style}
\end{figure}

\label{sec:intro}
Painting, an art form rich in diversity and complexity, has been an integral part of human culture throughout history. It encompasses a wide range of styles, from delicate watercolor scenes to intricate Chinese ink landscapes and detailed oil portraits. In recent decades, there has been a concerted effort to simulate these diverse artistic styles using non-photorealistic rendering techniques, including stroke-based and painterly rendering approaches~\cite{hertzmann1998painterly, winkenbach1996rendering}. While these methods have produced impressive results, they often rely on manual engineering, limiting their ability to create entirely novel styles.

Recent advances in machine learning have revolutionized image recognition and synthesis, opening up new possibilities for creative tasks such as painting. Machine learning techniques have been applied to various aspects of painting, including brush modeling~\cite{xie2012}, generating brush stroke paintings in specific artist styles~\cite{xie2015stroke}, and constructing stroke-based drawings~\cite{DBLP:journals/corr/HaE17}. Other approaches leverage generative adversarial networks \cite{goodfellow2014generative} and variational autoencoders~\cite{kingma2013auto} to emulate artistic styles \cite{zhu2017unpaired, zhou2018learning, DBLP:journals/corr/abs-1803-04469, karras2017progressive, sangkloy2017scribbler}.

In this paper, we focus on a more general and challenging problem of training a natural media painting agent from scratch using reinforcement learning methods. Our goal is to develop an agent that is able to perform a sequence of primitive drawing actions to produce a target output. Given a reference image, our painting agent aims to reproduce the identical or transformed version of that image in the simulated and real environment.

We present a novel automated painting framework that employs a painting agent trained through reinforcement learning for natural media painting. The primary objective of our painting agent is to faithfully reproduce a given reference image, either identically or in a transformed manner, in both simulated and real-world environments. In the simulated environment, our model can acquire complex painting policies through reinforcement learning. In the real environment, we have developed a method to transfer the learned policies while preserving their artistic capabilities.

The contributions of our work include:
\begin{itemize}
    \item The introduction of a novel deep reinforcement learning network meticulously designed for learning natural painting media within a simulated environment. Our approach exhibits the versatility to learn with or without human supervision and excels in navigating continuous high-dimensional action spaces, enabling it to effectively handle large and intricately detailed reference images.
    \item The development of an adaptive sim-to-real methodology tailored for deformable brushes. This methodology capitalizes on behavior cloning to initialize policies for painting tasks, facilitating the seamless transfer of learned policies from simulation to reality.
    \item A real painting environment featuring a robotic arm and brush, which corresponds to the MyPaint virtual environment. This real-world setup allows us to undertake complex artistic endeavors, including painting various subjects.
\end{itemize}

We have rigorously evaluated our results using a diverse set of reference images, spanning a wide range of artistic styles, as illustrated in Figure \ref{fig:style}. This evaluation encompassed both simulated and real robot setups. Our virtual painting agent exhibits the capability to generate high-resolution outputs tailored to different painting media. Concurrently, our real robot adeptly replicates the subtleties of these references across a spectrum of artistic styles. Through this implementation, we aim to provide a robust and practical solution for high-degree-of-freedom end-effector manipulation tasks. Our method is meticulously designed to discern and adapt to the intricate relationships between actions and environmental changes.

\section{Related Work}
\subsection{Learning-based Drawing}
There have been several attempts to address related problems in this domain. Xie et al.~\cite{xie2012,xie2015stroke,xie2013personal} proposed a series of works to simulate strokes using reinforcement learning and inverse reinforcement learning. These approaches learn a policy either from reward functions or expert demonstrations. Unlike our goal, Xie et al.~\cite{xie2012,xie2015stroke,xie2013personal} primarily focus on designing reward functions for generating oriental painting strokes, and their methods require expert demonstrations for supervision. Recently, Ha et al.~\cite{DBLP:journals/corr/HaE17} collected a large-scale dataset of millions of simple sketches of common objects with the corresponding recording of painting actions. Based on this dataset, a recurrent neural network model is trained in a supervised manner to encode and re-synthesize action sequences, and the trained model is shown to be capable of generating new sketches. Following \cite{DBLP:journals/corr/HaE17}, Zhou et al.~\cite{zhou2018learning} exploit reinforcement learning and imitation learning to reduce the amount of supervision needed to train such a sketch generation model. Distinct from \cite{DBLP:journals/corr/HaE17,zhou2018learning}, our painting agent operates in a complex SSPE with a continuous action space involving brush width and color, and our approach learns its policy network completely without human supervision.

\subsection{Visual Generative Methods}
Visual generative methods typically directly synthesize visual output in pixel spaces, which is fundamentally distinct from our approach. Image analogies by Hertzmann et al.~\cite{hertzmann2001image} solve this problem by introducing a non-parametric texture model. More recent approaches, based on CNNs and using large datasets of input-output training image pairs, learn the mapping function~\cite{gatys2015neural}. Inspired by the idea of variational autoencoders~\cite{kingma2013auto}, Johnson et al.~\cite{johnson2016perceptual} introduced the concept of perceptual loss to implement style transferring between paired datasets. Inspired by the idea of generative adversarial networks (GANs)~\cite{goodfellow2014generative}, Zhu et al.~\cite{zhu2017unpaired} learn the mapping without paired training examples using Cycle-Consistent Adversarial Networks. These methods have been successful at generating natural images~\cite{karras2017progressive,sangkloy2017scribbler}, artistic images~\cite{li2017universal}, and videos~\cite{vondrick2016generating,li2018flow}. In terms of the final rendering, current visual generative methods can produce results in various painting styles using a limited training dataset. However, compared to our method, these generative methods may fail to achieve high-resolution results. For the purpose of interactive artistic creation, the stroke-based approach can generate trajectories and intermediate painting states. Another advantage of the stroke-based method is that the final results are trajectories of the paintbrush, which can be deployed in different synthetic natural media painting environments and real painting environments using robot arms.

\subsection{Reinforcement Learning-based Painting Methods}
In the development of robotic painting algorithms, various approaches have been investigated. In significant work by Lee et al.~\cite{lee2022scratch}, a hierarchical reinforcement learning (RL) model was proposed for painting tasks, where a high-level controller learns the painting policy and a low-level manipulator adapts to the deformation of the brush. This dual-layered approach has been a critical reference point for our research. However, in our proposed method, we have prioritized efficiency and higher-dimensional control. Our model is capable of managing sophisticated control strategies, including the adjustment of pressure, stroke width, and depth.

Other studies, such as those by Chen et al.~\cite{chen2022gtgraffiti}, El et al.~\cite{el2019mobile}, and Vempati et al.~\cite{vempati2020data}, have focused on learning low-level manipulation policies to tackle challenges presented by uneven painting surfaces. We also incorporate these strategies into our method, illustrating its versatility and adaptability. A distinctive feature of our approach, compared to these studies, is that our method does not require explicit environmental modeling. Consequently, our algorithm exhibits broader applicability in real-world scenarios and a wider range of painting tasks, marking a significant contribution to the field of robotic painting algorithms.

\begin{table}[t]
\begin{tabular}{ll}
\hline  
Symbol & Meaning   \\
\hline
$t$ & step index \\
$s_t$ & current painting state of step $t$, canvas\\
$s^*$ & target painting state, reference image\\
$\hat{s^*}$ & reproduction of $s^*$ \\
$o_t$ & observation of step $t$ \\
$a_t$ & action of step $t$, $a_t=[ \alpha_t, l_t, w_t, c_t ]$\\
$r_t$ & reward of step $t$ \\
$q_t$ & accumulated reward of step $t$ \\
$\gamma$ & discount factor for computing the reward \\
$p_t$ & position of the paintbrush of step $t$ \\

\hline
$\pi$ & painting policy, predict $a$ by $o$\\
$V_\pi$ & value function of the painting policy, \\
& predict $r$ by $o$ \\

$R(a_t, s_t)$ & render function, render action to $s_t$\\
$O(s^*, s_t)$ & observation function, encode the current \\
 & state and the target state  \\
$L(s, s^*)$ & loss function, measuring distance between \\
    & state $s$ and objective state $s^*$\\
\hline
$\alpha_t$ & angle of action $a_t$\\
$l_t$ & length of action $a_t$\\
$w_t$ & stroke width of action $a_t$\\
$c_t$ & color descriptor of action $a_t$\\
\hline
\end{tabular}
\caption{\label{Fig:param} Notation Summary}
\end{table}

\begin{figure}[t]
\centering
\includegraphics[width=0.45\textwidth]{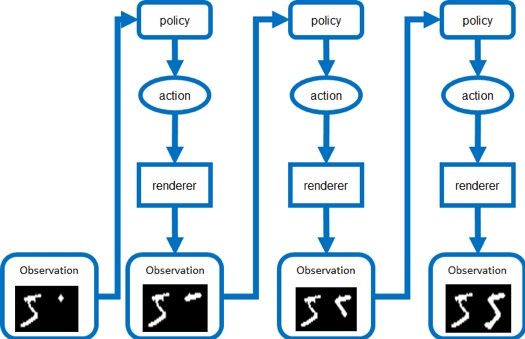}
\caption{{\em Overview of Training/Rollout Process:} For each time step, the current state of the canvas and the reference image form the observation for the policy network. Based on the observation, the policy network selects an action to execute and update the canvas accordingly.}
\label{fig:train}
\end{figure}

\begin{figure}
\centering
\includegraphics[width=0.48\textwidth]{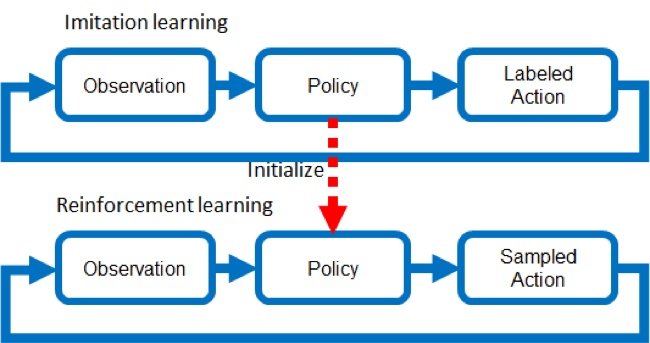}
\caption{\textit{Behavior Cloning for Policy Initialization:} We utilize a behavior cloning algorithm to train the policy, extending the action space to initialize the reinforcement learning (RL) policy within a real environment setup. The action space used in behavior cloning is a subspace of the RL action space and includes direction and on/off canvas actions. This initialization process bridges the gap between behavior cloning and RL, facilitating effective policy learning in the real environment.}
\label{fig:imitate}
\end{figure}
\section{Training a Painting Policy}
\label{sec:rl}

In this section, we delve into the technical details of our painting agent based on reinforcement learning. We begin by introducing the fundamental components of reinforcement learning, encompassing the action space, observation, reward, and policy network. Subsequently, we elucidate the intricacies of our training and runtime algorithms, along with methodologies aimed at enhancing learning efficiency, including curriculum learning, difficulty-based sampling, and self-supervised learning.

\subsection{Policy Representation}
The policy of our painting agent encompasses the definition of actions, observations, rewards, and the architecture of the policy network. The action space characterizes the degrees of freedom of the painting agent, representing the output of the policy network. Observations capture the state of the painting process, serving as input to the policy network. The reward function quantifies the effectiveness of painting actions in achieving the desired configuration, as determined by the environment. The policy network's structure dictates the technical implementation of the machine learning approach.

\begin{algorithm}[t]
  \caption{Rollout Algorithm}
  \label{alg:test}
  \begin{algorithmic}[1]
    \REQUIRE Reference image $s^*$ with size $(h_{s^*}, w_{s^*})$, the learned painting policy $\pi$ with observation size $(h_o, w_o)$
    \ENSURE Final rendering $\hat{s^*}$
    
    \WHILE {$||I-I^*|| > Thresh_{sim}$}
        \STATE $h = \text{rand}(h_{s^*})$ // Sample a 2-dimensional point within the image to start the stroke
        \STATE $w = \text{rand}(w_{s^*})$
        \STATE $o = s[h-\frac{h_o}{2}:h+\frac{h_o}{2}, w-\frac{w_o}{2}:w+\frac{w_o}{2}]$ // Get observation
        \STATE $r = 1$ // Initialize the predicted reward
        \WHILE {$r > 0$}
            \STATE $a = \pi(o)$ // Predict the painting action
            \STATE $r = V_\pi(o)$ // Predict the expected reward
            \STATE $s = R(s, a)$ // Render the action
            \STATE $h = h + l \times \cos(\alpha)$ // Update the stroke position
            \STATE $w = w + w \times \sin(\alpha)$ 
            \STATE $o = s[h-\frac{h_o}{2}:h+\frac{h_o}{2}, w-\frac{w_o}{2}:w+\frac{w_o}{2}]$ // Update the observation
        \ENDWHILE
    \ENDWHILE
    \RETURN $s$
  \end{algorithmic}
\end{algorithm}

\subsubsection{Action Space}
\label{sec:action}
To capture the essence of painting behavior, we represent actions using stroke properties, including angle, length, size, and color. Specifically, we define the action as a 6-dimensional vector, $a_t=[\alpha_t, l_t, w_t, c_{rt}, c_{gt}, c_{bt}] \in \mathbb{R}^6$, with each value normalized to $[0, 1]$. The action space is continuous, enabling us to employ policy gradient-based reinforcement learning algorithms. Notably, when $w=0$, the brush moves above the canvas without applying paint.

\subsubsection{Observation}
\label{sec:obs}
Our approach extends the observation $o_t$ of the painting state to encompass the reference image $s^*$ as part of the observation, defined as $o_t = \{s_t, p_t\}$. This inclusion enables the model's generalization across different reference images. In all our experiments, both the reference image and the canvas are encoded as observations, representing the current state and the goal state of the agent.

We tackle the challenge of incorporating positional information by adopting an egocentric observation strategy. In this strategy, the paintbrush remains centered on the canvas, with the canvas and reference image adjusted accordingly. This approach simplifies the action space, eliminates the need for a replay buffer, and renders training in a continuous action space and large state space feasible. The state observation $o_t$ is defined in Eq. \ref{eq:obs}, where $(h_p, w_p)$ denote the 2D position of the paintbrush, and $(h_o, w_o)$ represent the size of the egocentric window.

\begin{equation}
\label{eq:obs}
\begin{split}
    o_t = &\left\{ s_t\left[h_p-\frac{h_o}{2}:h_p+\frac{h_o}{2}, w_p-\frac{w_o}{2}:w_p+\frac{w_o}{2}\right]\right., \\ &\left.s^*\left[h_p-\frac{h_o}{2}:h_p+\frac{h_o}{2}, w_p-\frac{w_o}{2}:w_p+\frac{w_o}{2}\right] \right\}.
\end{split}
\end{equation}

This definition of observation allows us to incorporate the paintbrush's position and enables the generalization of training data.

We illustrate our rollout algorithm in Algorithm \ref{alg:test}.

\subsubsection{Reward}
\label{sec:reward}
In our setup, the reward for each action is determined by the difference between the canvas and the reference image. A loss function is employed to calculate the action's reward during each reinforcement learning iteration. To incentivize the painting agent to match the color and shape of the reference image precisely rather than aiming for an average color, we slightly modify the $L_2$ loss into $L_{\frac{1}{2}}$,

\begin{equation}
L_{\frac{1}{2}}(s, s^*) = \frac{\sum^h_{i=1}\sum^w_{j=1}\sum^c_{k=1}|s_{ijk}-s^*_{ijk}|^{\frac{1}{2}}}{hwc},
\end{equation}

where the image $s$ and the reference image $s^*$ are matrices with dimensions $h \times w \times c$. Here, $w$ and $h$ denote the width and height of the image, while $c$ represents the number of color channels.

After defining the loss between $I$ and $I^{ref}$, we normalize $r_t$ using Eq. \ref{eq:reward}, such that $r_t \in (-\infty, 1]$.

\begin{equation}
\label{eq:reward}
r_t = \frac{L(s_{t-1}, s^*) - L(s_t, s^*)}{ L(s_0, s^*)}
\end{equation}

\subsubsection{Policy Network}
The first hidden layer applies convolution with 64 $8 \times 8$ filters and a stride of 4. The second layer employs convolution with 64 $4 \times 4$ filters and a stride of 2, followed by the third layer using convolution with 64 $3 \times 3$ filters and a stride of 1. Subsequently, the network connects to a fully-connected layer comprising 512 neurons. All layers employ the ReLU activation function \cite{krizhevsky2012imagenet}.

\subsubsection{Curriculum Learning}
\label{sec:curriculum}
Given the continuous action space $a \in \mathbb{R}^6$, the sampling space can grow significantly as the number of time steps increases. Moreover, policy gradient-based reinforcement learning algorithms may introduce noise that overwhelms the signal. To efficiently train the model, we adopt a curriculum learning approach, wherein the number of sampled trajectories increases during training episodes. Consequently, the agent can learn policies incrementally and generate relatively long strokes compared to models trained without this technique. The agent tends to seek rewards greedily within the limited time steps.

Another primary challenge arises from the bias among different samples. In conventional RL tasks, the goal is typically fixed. In our case, however, the reference image must change to prevent overfitting. To overcome this challenge, we implement difficulty-based data sampling. In reinforcement learning, the optimal policy $\pi^*$ maximizes the expected long-term reward $q_t$, which accumulates rewards $r_t$ over a time horizon $t_{\max}$ of steps, incorporating a discount factor $\gamma \in \mathbb{R}$,

\begin{equation}
q_t = \sum^{t_{\max}}_{t=1}{r_t\gamma^t},
\end{equation}

where $t_{\max} \in \mathbb{Z}$ represents the maximum number of steps for each trial.

For a painting policy, numerous goal configurations are sparsely distributed across a high-dimensional space, posing challenges for the convergence of the agent's learning process. We adapt the horizon parameter $t_{\max}$ by introducing a reward threshold $r_{\mbox{\scriptsize thresh}}$ and gradually increasing it during training as:

\begin{equation}
\hat{t}_{\max} = \argmin_i(r_i>r_{\mbox{\scriptsize thresh}}).
\end{equation}

With this redefined horizon parameter, the policy gradient algorithm can efficiently converge when dealing with a set of complex goal configurations. This encourages the policy to seek rewards greedily within limited time steps, thus reducing the exploration space.

\section{Sim-to-Real Brush Manipulation}

In this section, we will provide a detailed explanation of the methods employed for sim2real transfer from the painting policy in Section \ref{sec:rl}. The objective is to seamlessly transfer the painting policy learned in simulation to real-world robotic drawing tasks. This transfer is essential for achieving high-quality brush manipulation and stroke control in real-world scenarios.

To effectively control the shape of strokes and ensure precise interactions between the brush and various painting media, such as ink, water, and foam, it is imperative to estimate pressure accurately. Pressure plays a pivotal role in determining the thickness and texture of strokes, significantly impacting the quality of artwork produced by the robot.

Unlike traditional methods that rely on force sensors, our approach leverages advanced modeling and image analysis techniques to estimate pressure, making it suitable for a wide range of applications where force sensing may not be feasible.

In our practical experiments, we adopted a hybrid approach, as outlined in Fig. \ref{fig:imitate}. Initially, we utilized flexible end-effector image capture to determine the optimal pressure range. Subsequently, we employed a stroke image sampling technique to establish a precise mapping.

Regarding the policy we have acquired, it can be deconstructed into two distinct components: the high-level and low-level policies.
The high-level policy is trained through behavior cloning, enhancing the standardization of stroke order, particularly in the context of handwriting.
In contrast, the low-level policy is developed using an efficient sampling-based reinforcement learning methodology. This policy functions as a mapping mechanism, translating the original reinforcement learning low-level policy into tangible actions within the real-world environment.

\subsection{Contact Force Estimation}
Accurately estimating the contact force between the pen tip and the painting media is a crucial aspect of robotic brush manipulation. However, precise force sensors are often unavailable. Therefore, we employ image analysis methods to infer pressure values.

\begin{figure}[t]
\centering
\begin{subfigure}{0.24\textwidth}
  \centering
  \includegraphics[width=\linewidth]{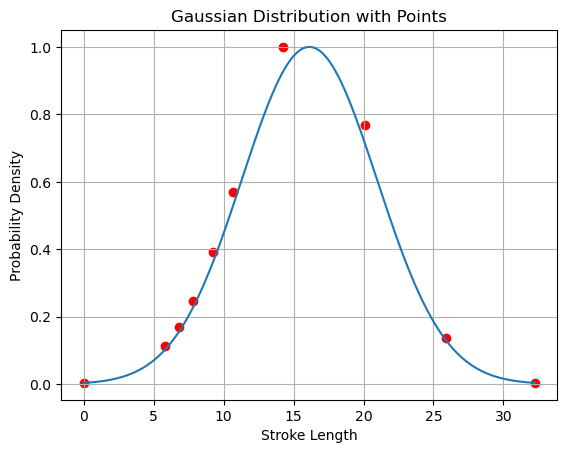}
  \caption{Mean: 0.5, Standard Deviation: 0.3.}
  \label{subfig:gaussian-a}
\end{subfigure}\hfill
\begin{subfigure}{0.24\textwidth}
  \centering
  \includegraphics[width=\linewidth]{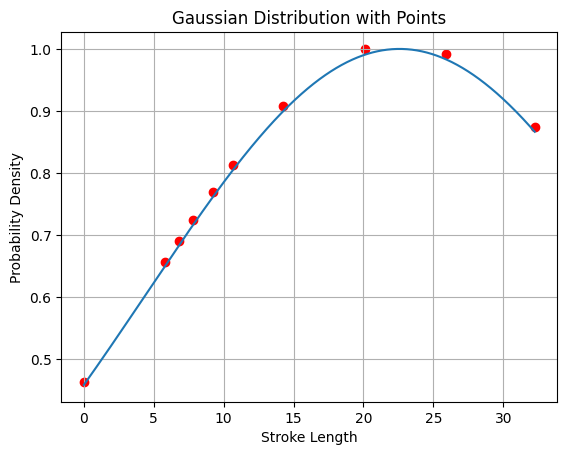}
  \caption{Mean: 0.7, Standard Deviation: 0.8.}
  \label{subfig:gaussian-b}
\end{subfigure}
\caption{\textit{Effect of Gaussian Stroke Model on Stylization:} We model a long stroke composed of segments using a Gaussian distribution. These correspond to the first column in Fig. \ref{fig:style}, where variations in artistic style are achieved by adjusting the Gaussian parameters.}
\label{fig:gaussian}
\end{figure}

\subsubsection{Observation of Stroke Images}
This approach involves indirectly observing environmental changes, specifically the stroke images on the paper, to infer variations in pressure. It is an intuitive method where we record the shape of strokes and the configuration of the robotic arm. We can then interpolate to obtain the desired stroke characteristics.

However, finding a suitable arm configuration is not straightforward. Similar to training reinforcement learning (RL) in simulation, this method requires extensive sampling, with many instances yielding no positive rewards due to the limited deformation range of the brush.

\subsubsection{Observation of End-Effector Images}
In contrast to observing stroke images, this method offers a more direct approach. It involves capturing the shape changes of the flexible end effector.

While this method may be susceptible to image noise, it provides valuable information about the pressure limit of the flexible object. We utilize linear fitting to identify the point at which deformation no longer occurs, treating it as the pressure limit.

\begin{algorithm}[t]
  \caption{Hybrid Pressure Estimation (Recursive)}
  \label{alg:hybrid_pressure_estimation_recursive}
  \begin{algorithmic}[1]
    \REQUIRE Initial robot action in C-space for pressure [$a_{\text{p\_max}}$, $a_{\text{p\_min}}$], real robot renderer $s = R(a)$
    \ENSURE Optimal maximum pressure $a^*_{\text{p\_max}}$, minimum pressure $a^*_{\text{p\_min}}$, mapping function from stroke to the real robot configuration $a' = M(s)$
    
    \STATE $\textbf{function} \, \text{EstimatePressure}(a_{\text{p\_max}}, a_{\text{p\_min}})$
    
    \IF {$(a_{\text{p\_max}} - a_{\text{p\_min}}) \leq \text{a\_step}$}
        \RETURN $M(s)$
    \ELSE
        \STATE $a_{\text{p\_guess}} \gets (a_{\text{p\_min}} + a_{\text{p\_max}}) / 2$
        \STATE $s \gets R(a_{\text{p\_guess}})$
        
        \STATE \textbf{update policy} $M$ with $s$ and $a_{\text{p\_guess}}$
        
        \RETURN \text{EstimatePressure}($a_{\text{p\_max}}$, $a_{\text{p\_guess}}$)
        \RETURN \text{EstimatePressure}($a_{\text{p\_guess}}$, $a_{\text{p\_min}}$)
    \ENDIF
    
    \STATE $\textbf{end function}$
    \STATE $M(s) \gets \text{EstimatePressure}(a_{\text{p\_max}}, a_{\text{p\_min}})$
    
    \RETURN  $M(s)$
  \end{algorithmic}
\end{algorithm}

\subsection{Mapping Actions from Simulation to Reality}

In Section \ref{sec:rl}, we defined actions in a simulated environment, which may differ from the actions required in the real-world environment. Therefore, we need to map robot actions from the simulated environment's action space to the robot's configuration space in the real world.

The first challenge is that the painting plane in the simulated environment differs from the real robot environment. Therefore, we need to find a 2D plane in the 3D configuration space to serve as the painting space. The action mapping formula is computed similarly to the camera's extrinsic calibration.

The second challenge arises because certain actions cannot be directly translated into robot movements but still have a limited visual effect. These include:
\begin{enumerate}
    \item Stroke thickness, which can only be adjusted by changing the brush's contact force.
    \item Color, which, in our setup, is limited to monochrome. Color changes are achieved through interactions with the environment, such as dipping in ink, water, or interacting with a sponge.
    \item Tilt, which our 3-DoF robot cannot directly achieve due to limited kinematics.
\end{enumerate}
To approximate these effects, we employ the following methods:

\subsubsection{Gaussian Modeling of Strokes}
The key to achieving artistic font treatment is to emulate the stroke characteristics of human artists. To accomplish this, we use Gaussian modeling for each stroke. This model captures the distribution of the stroke's centroid and pressure, allowing us to generate artistic fonts with various styles. Fine-tuning these parameters enables us to create different types and styles of strokes, achieving font diversity.

\subsubsection{2D to 3D Action Projection}

To match the actions from the simulated environment to the real robot's configuration space, we need to project 2D actions into a 3D configuration space. This projection can be defined using the following equation, which is similar to a camera's extrinsic calibration projection:

\[
\begin{bmatrix}
x_{\text{robot}} \\
y_{\text{robot}} \\
z_{\text{robot}}
\end{bmatrix}
=
\begin{bmatrix}
R & T \\
0 & 1
\end{bmatrix}
\begin{bmatrix}
x_{\text{painting}} \\
y_{\text{painting}} \\
1
\end{bmatrix}
\]

Here, \(x_{\text{robot}}\), \(y_{\text{robot}}\) and \(z_{\text{robot}}\) represent the robot's coordinates. \(x_{\text{painting}}\) and \(y_{\text{painting}}\) are the desired painting coordinates in 2D space. The transformation matrix \(\begin{bmatrix}R & T \\ 0 & 1\end{bmatrix}\) maps the 2D painting coordinates to the 3D robot configuration, allowing us to generate actions that correspond to the desired painting locations and orientations in the real world.

\section{Behavior Cloning}
\label{sec:behavior_cloning}

Behavior cloning leverages a paired dataset comprising observations and corresponding actions to train a policy to mimic expert trajectories or behaviors. In our context, the expert trajectory is encoded in the paired dataset $\{o_{(t)}, a_{(t)}\}$. We employ behavior cloning to initialize the policy network for reinforcement learning, using the supervised policy trained with the paired data. The paired dataset can be generated by a human expert or an optimal algorithm with global knowledge, which our painting agent lacks. Once we obtain the paired dataset $\{o_{(t)}, a_{(t)}\}$, one common approach is to apply supervised learning based on regression or classification to train the policy. The training process can be formulated as an optimization problem:

\begin{equation}
\label{eq:problem}
\pi^* = \argmin\sum_{t}^{N}{||\pi(o_t) - a_t ||}.
\end{equation}

Generating an expert dataset for our painting application can be challenging due to the significant variation in reference images and painting actions. However, we can create a paired dataset by rolling out a policy during the RL training process. Additionally, there are existing datasets like KanjiVG and Google's Quick, Draw! that provide paired supervised data \cite{kanjivg, quickdraw}.

\section{Experiment}

\subsection{Setup}
For our simulated painting setup, we created an environment that allows the painting agent to explore a high-dimensional action space and observation space based on MyPaint \cite{mypaint}. 

For the real brush manipulation experiment, we implement our approach using an UltraArm, which features 3 DoFs for movement as shown in Fig. \ref{fig:style}. The primary experimental setup includes a water pot and foam, allowing the robot to manipulate a paintbrush by absorbing water, squeezing it, or using the object to reshape it. This setup serves to demonstrate that our method can effectively learn the complexity of high DoF end-effector manipulation tasks in a practical and realistic scenario.

By incorporating the water pot and foam into the experimental setup, we introduce additional challenges that the robot must learn to overcome. These include controlling the amount of water absorbed by the paintbrush, adjusting the pressure applied when squeezing or reshaping the brush, and maintaining a stable grip on the brush throughout the manipulation process. These added complexities showcase the adaptability and effectiveness of our approach in handling diverse manipulation tasks involving deformable materials and intricate interactions with the environment.

\subsection{Data Preparation}
\begin{figure}[t]
  \centering
    \includegraphics[width=0.45\textwidth]{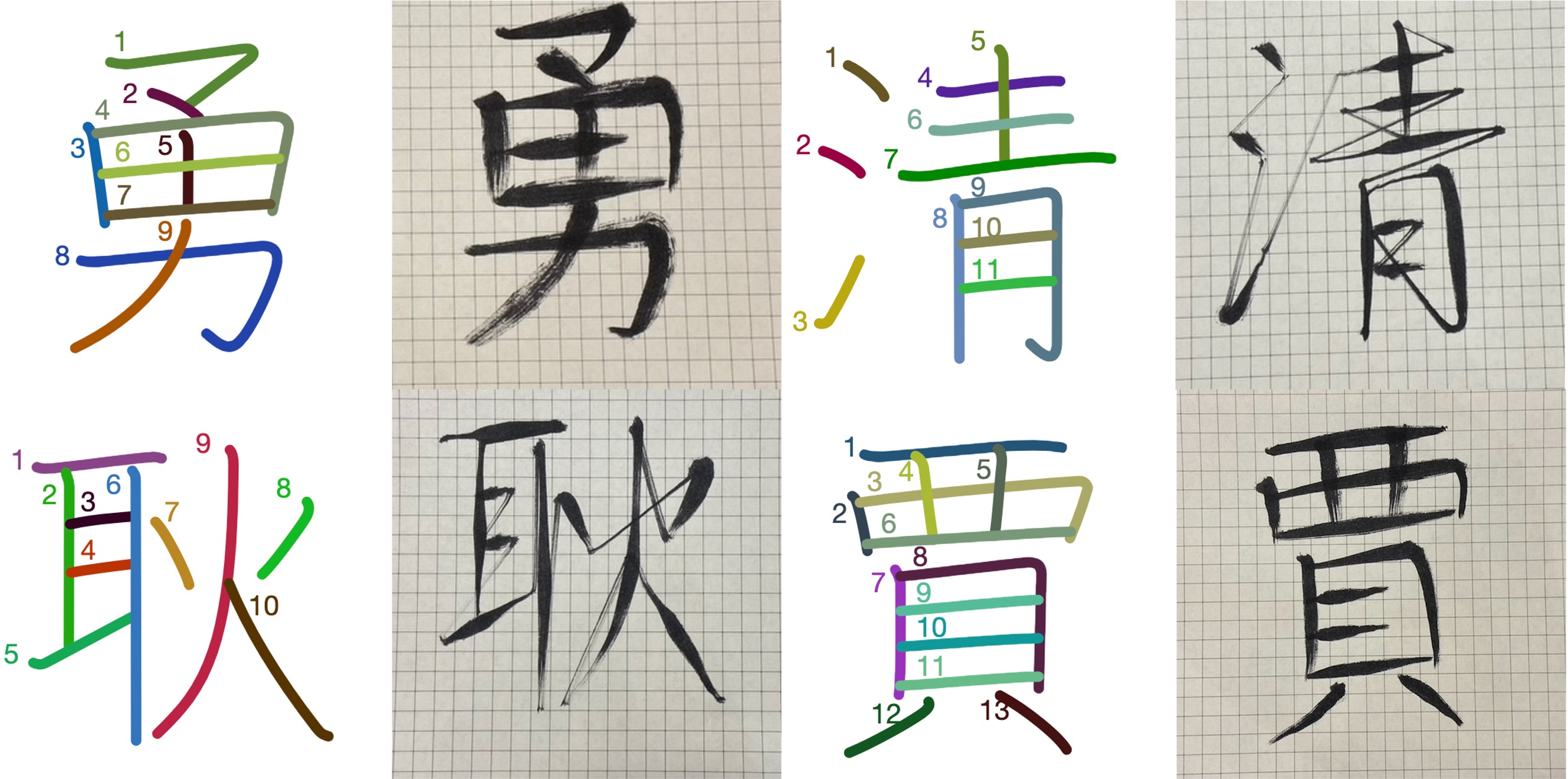}    
    \caption{Illustration of KanjiVG's labeled data used for behavior cloning (Column 1,3) and a result generated by a 3-DoF robot (Column 2,4).}
  \label{fig:data}
\end{figure}

In the scope of our real-robot experiments, we selected the KanjiVG dataset \cite{kanjivg} for our training endeavors. This dataset, rich in its depth, provides detailed stroke information for approximately 2,000 distinct characters. Every individual character within the dataset has been complemented with associated painting actions, which are vividly depicted in Fig.~\ref{fig:data}, columns 1 and 3. This dataset, having been meticulously collated from human participants, establishes itself as a premier choice when leveraging behavior cloning in the domain of robotic calligraphy.

Within the framework of our reinforcement learning (RL) strategy, we leaned on the acclaimed CelebA dataset \cite{liu2015celeba} to facilitate the training of our painting agent. It's important to note that our rollout algorithm was architectured employing MyPaint \cite{mypaint}, a decision made to ensure the results seamlessly mirror the characteristics of natural media. The nuances of the painting model are distilled implicitly, rooted in the foundational knowledge embedded in the environment model. The versatility and robustness of our algorithm are showcased in Fig.~\ref{fig:style}.

\subsection{Evaluation}
We demonstrated the advantages of our approach by computing performance and comparing visual effects. We designed three experiments to evaluate the performance of our algorithms.

\begin{figure}[ht]
\centering
\includegraphics[width=0.4\textwidth]{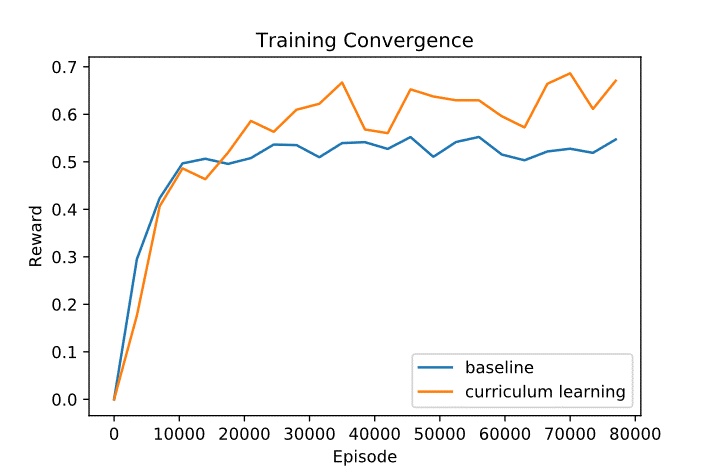}
\caption{\emph{Curriculum Learning} This figure compares the learning curve between the approach using curriculum learning and the baseline. The y-axis denotes the average rewards of the trained model in a validation dataset, and the x-axis denotes the training episodes. Both approaches converged after a certain number of steps, but the approach with curriculum learning performed better with a higher reward value. The total training steps used in both approaches are about $10^6$.}
\label{fig:learning_curve}
\end{figure}

For the first experiment, we computed the learning curve of the baseline model and the model with curriculum learning (Sec.~\ref{sec:curriculum}), as shown in Fig.~\ref{fig:learning_curve}. Both models converged within $78,000$ episodes. The y-axis denotes the average rewards of the trained model in a validation dataset, and the x-axis denotes the training episodes. As the training process proceeded, the average rewards grew, showing that curriculum learning can improve the reinforcement learning to converge to a better policy.

For the second experiment, we evaluated the performance of the high-resolution reference images. We computed the $L_2$ loss and cumulative rewards  and compared our approach with behavior cloning, reinforcement learning, and a combined. We drew $1000$ $400 \times 400$ patches from 10 reference images to construct the benchmark. Moreover, we iteratively applied both algorithms $1000$ times to reproduce the reference images. We used the same training dataset with images to train the models. As shown in Table~\ref{tab:rewards}, self-supervised learning had a lower $L_2$ loss, although both methods performed well in terms of cumulative rewards.

\begin{table}[t]
\setlength{\tabcolsep}{3pt}
\centering
\begin{tabular}{|l|c|c|}
\hline
Approaches & Cumulative Rewards & $L_2$ Loss \\
\hline
Behavior Cloning & $20.15$ & $512$ \\
\hline
Reinforcement Learning & $97.74$ & $1920$ \\
\hline
Our Combined Scheme & $98.25$ & $1485$ \\
\hline 
\end{tabular}
\caption{\emph{Evaluation of Painting Approaches} We evaluated the performance of behavior cloning, reinforcement learning, and our combined scheme by computing the average cumulative reward and $L_2$ loss between the final rendering and the reference image on the test dataset.}
\label{tab:rewards}
\end{table}

\section{Conclusion, Limitations, and Future Work}
In this study, we introduced an innovative approach for training a reinforced natural media painting agent, designed specifically for stroke-based image reproduction. Leveraging a novel reinforcement learning framework, we entered the domain of high-dimensional and continuous action spaces.

While our approach demonstrates substantial promise, it's important to acknowledge its limitations. A significant constraint lies in the policy's dependence on the training data. Despite reinforcement learning's inherent generalization capabilities, the policy's effectiveness is closely tied to the distribution of training data, potentially limiting its performance on significantly different unseen data.

The real-world deployment of our painting agent is crucial. In the context of real robots, our agent showcases its artistic capabilities, emphasizing the importance of sim-to-real transfer as it seamlessly transitions from simulated training to real-world application.

For future research directions, we aim to expand the temporal horizon and action space within the painting environment, especially in challenging real-world settings. Additionally, while our current framework covers common stroke parameters like angle, length, brush size, and color, there's untapped potential in incorporating additional painting parameters, such as pen tilting, pen rotation, and pressure, into our policy framework. Exploring these dimensions promises to further enrich our painting agent's expressive capabilities, pushing the boundaries of robotic artistic creation.

In conclusion, our work represents progress in the realm of reinforcement learning and sim-to-real policy transfer for robotic manipulation. By addressing limitations and embarking on future research endeavors, we aim to unlock new frontiers at the intersection of art and artificial intelligence, ultimately enhancing the creative potential of machines in both simulated and real-world contexts.

\bibliographystyle{IEEEtran}
\bibliography{main}

\end{document}